%% file: main.tex
\documentclass[conference]{IEEEtran}
\IEEEoverridecommandlockouts

\usepackage{cite}
\usepackage{amsmath,amssymb,amsfonts}
\usepackage{algorithmic}
\usepackage{graphicx}
\usepackage{textcomp}
\usepackage{xcolor}
\usepackage{amsmath}
\usepackage{balance}

\usepackage{booktabs}
\def\BibTeX{{\rm B\kern-.05em{\sc i\kern-.025em b}\kern-.08em
    T\kern-.1667em\lower.7ex\hbox{E}\kern-.125emX}}
\begin{document}

\title{MAGE-ID: A Multimodal Generative Framework for Intrusion Detection Systems}


\author{
Mahdi Arab Loodaricheh, Mohammad Hossein Manshaei, and Anita Raja\\
Department of Computer Science, Hunter College and The Graduate Center,\\ City University of New York, NY, USA\\
marabloodaricheh@gradcenter.cuny.edu, \{mohammad.manshaei, anita.raja\}@hunter.cuny.edu
}

\maketitle

\begin{abstract}
Modern Intrusion Detection Systems (IDS) face severe challenges due to heterogeneous network traffic, evolving cyber threats, and pronounced data imbalance between benign and attack flows. While generative models have shown promise in data augmentation, existing approaches are limited to single modalities and fail to capture cross-domain dependencies. This paper introduces MAGE-ID (Multimodal Attack Generator for Intrusion Detection), a diffusion-based generative framework that couples tabular flow features with their transformed images through a unified latent prior. By jointly training Transformer- and CNN-based variational encoders with an EDM-style denoiser, MAGE-ID achieves balanced and coherent multimodal synthesis. Evaluations on CIC-IDS-2017 and NSL-KDD demonstrate significant improvements in fidelity, diversity, and downstream detection performance over TabSyn and TabDDPM, highlighting MAGE-ID’s effectiveness for multimodal IDS augmentation.
\end{abstract}

\begin{IEEEkeywords}
Intrusion Detection Systems, Generative Models, Diffusion Models, Multimodal Learning, Tabular Data Synthesis.
\end{IEEEkeywords}

\input{intro}
\input{related}
\input{method}

\input{experiments}
\input{results}
\input{conclusion}
\bibliographystyle{IEEEtran}
\bibliography{biblio}

\end{document}

%% file: intro.tex
\section{Introduction}
\label{sec:intro}

Modern enterprise and IoT networks produce large volumes of heterogeneous, high-dimensional traffic and face increasingly sophisticated cyber threats. \textit{Intrusion Detection Systems (IDS)} remain essential but struggle to adapt: signature-based IDS detect only known attacks \cite{sommer2010outside}, while anomaly-based methods can identify unseen threats yet often suffer from high false positives and limited adaptability \cite{buczak2016survey}. This underscores the need for adaptive IDS frameworks that generalize beyond static signatures to detect emerging attack variants.


Beyond data scarcity, another key limitation of current IDS research is its reliance on \textit{single-modality analysis}. In practice, cyber-attacks manifest across multiple heterogeneous sources: network flow statistics (tabular data), temporal patterns in packet sequences (time-series), and image-like structures. Focusing on only one of these modalities risks missing critical contextual information. Recent multimodal IDS studies demonstrate that fusing heterogeneous data sources (e.g., network flows with payload bytes, or IoT traffic with textual features) significantly improves detection accuracy compared to unimodal baselines \cite{kiflay2024multimodal,ullah2024enhanced}. However, integrating multimodal data into generative augmentation remains largely unexplored. Existing generative IDS frameworks typically target a single modality, such as tabular network flows, and do not ensure consistency across multiple data types. This creates a gap: without coordinated multimodal synthesis, synthetic data cannot fully represent the cross-domain nature of modern cyber threats~\cite{yu2025latentdiffusion}.

\begin{figure*}[h]
  \centering
  \includegraphics[width=0.87\textwidth]{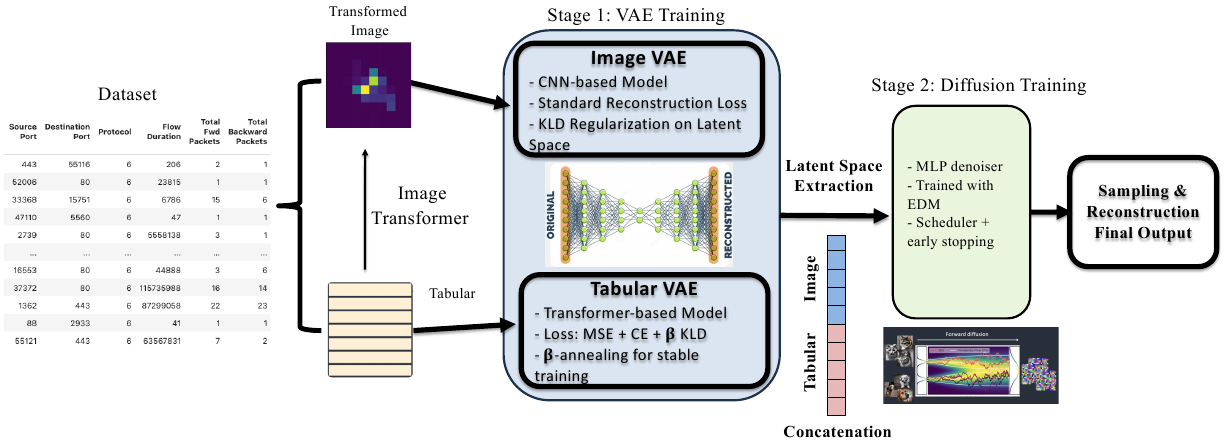}
   \caption{Overview of the proposed \textbf{MAGE-ID} framework. Stage~1 trains Transformer- and CNN-based VAEs on tabular and transformed images, while Stage~2 models a joint latent space with an EDM-style denoiser for coherent multimodal generation.}

  \label{fig:mageid_pipeline}
\end{figure*}

To close these gaps, we present \textbf{MAGE-ID} (\textit{Multimodal Attack Generator for Intrusion Detection}), a two-stage \emph{multimodal} generator that couples tabular flow records with their transformed images through a unified latent diffusion prior. Modality-specific Variational Autoencoders (VAEs)—a Transformer-based VAE for tabular data and a CNN-based VAE for images—learn reconstruction-friendly posteriors whose standardized latents are concatenated and modeled by an EDM-style (Elucidated Diffusion Model) denoiser capturing cross-modal dependencies and noise-conditioned dynamics. This design targets scarce attack types by enabling balanced sampling while preserving fidelity and support coverage. Our contributions are threefold: (i) a unified latent-diffusion formulation enforcing coherence between tabular features and their transformed images; (ii) a stable training recipe—$\beta$-annealed VAEs with a noise-conditioned EDM denoiser—that enhances fidelity and diversity; and (iii) a comprehensive evaluation using PRDC (Precision, Recall, Density, and Coverage), SDMetrics Detectability, and Machine Learning Efficacy (MLE). Across CIC-IDS-2017~\cite{sharafaldin2018toward} and NSL-KDD~\cite{tavallaee2009detailed}, MAGE-ID surpasses recent diffusion baselines in fidelity, diversity, and downstream utility, establishing a compact yet effective framework for multimodal generative augmentation in intrusion.

In contrast to \textit{GANs}-based IDS augmentation methods that oversample rare attacks~\cite{wgan_dl_ids2024} or hybrid \textit{VAE–GAN} schemes to improve realism~\cite{tian2024vaewacgan}, MAGE-ID leverages recent advances in \textit{diffusion models}~\cite{kim2025diffusion} and multimodality. This diffusion-based latent prior overcomes instability and mode collapse, producing balanced, high-fidelity tabular–image samples that enhance minority-class coverage and multimodal consistency.

The rest of the paper is organized as follows: Section~\ref{sec:related} reviews related work; Section~\ref{sec:methodology} presents the MAGE-ID framework; Section~\ref{sec:experiments} describes the experimental setup; Section~\ref{sec:results} reports the results; and Section~\ref{sec:Conclusion} concludes the paper.

%% file: related.tex
\section{Related Work}
\label{sec:related}
Generative models have been increasingly explored to mitigate class imbalance~\cite{yousefi2025advar, loodaricheh2025handling} and enhance IDS robustness. GAN-based frameworks such as WGAN-DL-IDS~\cite{wgan_dl_ids2024} synthesize minority-class attack flows to improve recall, while diffusion-based generators~\cite{yu2025latentdiffusion} offer stable, high-fidelity synthesis of realistic and privacy-preserving network traffic. However, existing approaches remain \emph{unimodal}, relying solely on tabular features and ignoring cross-modal dependencies. Recent multimodal IDS architectures~\cite{kiflay2024multimodal,ullah2024enhanced} fuse heterogeneous data sources such as flow, payload, and byte-stream images to improve detection accuracy, but they are purely \emph{discriminative} and cannot generate coherent multimodal samples or address data imbalance. 
In contrast, the proposed \textbf{MAGE-ID} unifies generative and multimodal learning by capturing joint tabular–image representations in a shared latent space.

In addition, Diffusion-based tabular synthesis methods have demonstrated strong realism in structured data domains. \textit{TabDDPM}~\cite{kotelnikov2023tabddpm} models both continuous and categorical attributes through denoising diffusion, outperforming GAN and VAE baselines, while \textit{TabSyn}~\cite{zhang2024mixedtype} extends this approach with latent-space diffusion, coupling a VAE encoder with a score-based denoiser to capture complex feature dependencies. Parallel to this, the \textit{DeepInsight} framework~\cite{sharma2019deepinsight} transforms tabular features into structured 2D images by projecting correlated attributes into spatially adjacent pixels, later adapted for network traffic analysis in MAGNETO~\cite{andresini2021gan}. However, prior work treats DeepInsight purely as a discriminative representation tool. \textbf{MAGE-ID} couples tabular and DeepInsight image latents under a shared diffusion prior, enabling coherent multimodal generation with balanced fidelity and diversity.


%% file: method.tex
\section{Our Proposed Method: MAGE-ID}
\label{sec:methodology}

The proposed \textbf{MAGE-ID} framework is designed to address the challenges of modern intrusion detection, where heterogeneous network traffic exhibits strong feature dependencies and severe class imbalance between benign and attack flows. 
To enable multimodal synthesis, raw tabular flow records are transformed into structured \textit{DeepInsight} images~\cite{sharma2019deepinsight} that preserve feature-space correlations in a two-dimensional spatial layout, producing paired tabular and image representations for each network instance. 
As illustrated in Fig.~\ref{fig:mageid_pipeline}, MAGE-ID employs a two-stage generative architecture that jointly models these heterogeneous modalities in a unified latent space. 
The model learns to capture cross-modal dependencies between tabular statistics (e.g., packet-level and flow features) and their corresponding DeepInsight-transformed images, allowing it to synthesize realistic and distributionally consistent multimodal intrusion samples. 
By enriching under-represented attack categories such as DDoS and PortScan through balanced generative augmentation, the framework mitigates data imbalance while improving detection robustness and diversity in downstream IDS classifiers.

\textbf{Stage~1: Variational Encoding.}
Two modality-specific VAEs are jointly trained to learn latent representations. 
The \textit{Tabular VAE} employs a transformer-based encoder with a learnable \texttt{[CLS]} token that aggregates global contextual information across tabular features. 
The encoder outputs the mean ($\mu$) and log-variance ($\log \sigma^{2}$) of a latent Gaussian distribution, from which samples are drawn via the reparameterization trick ~\cite{kingma2013autoencoding}. 
The training objective combines \textit{Mean-Squared Error (MSE)} for numerical features, \textit{Cross-Entropy (CE)} for categorical features, and a $\beta$-weighted \textit{Kullback--Leibler (KL)} divergence term. 
The coefficient $\beta$ is \emph{annealed downward} during training to maintain a balance between reconstruction fidelity and latent regularization. 
The \textit{Image VAE}, built upon Convolutional Neural Networks (CNNs), encodes corresponding DeepInsight-transformed images into a compact latent space using a similar MSE~+~KL objective. 
After convergence, we draw $K$ posterior samples ($K{=}3$ in our experiments) from each VAE by reparameterization, discard the tabular \texttt{[CLS]} token, and concatenate the resulting tabular and image latents to form an \emph{aggregated multimodal posterior}. We chose $K{=}3$ because it provides sufficient latent variability to stabilize multimodal diffusion training while avoiding unnecessary computational overhead observed with larger $K$ values.
Each latent dimension is then standardized as $(z-\mu)/\sigma$ prior to diffusion training, with the transform inverted after sampling.

\textbf{Stage~2: Diffusion-Based Generation.}
The standardized multimodal latent space is modeled with an \textit{EDM-style denoiser} implemented as a Multi-Layer Perceptron (MLP). 
The denoiser learns to reconstruct clean latents from noised inputs using $\sigma$-dependent preconditioning and weighting, effectively capturing the joint distribution of tabular and image representations. 
During synthesis, Gaussian noise is iteratively denoised using an EDM sampler with mild stochasticity (\(S_{\text{noise}}{=}1.2\)) to enhance coverage and reduce mode collapse. 
The resulting latents are unwhitened and decoded through the trained VAEs to produce synthetic tabular rows and corresponding images. 
The overall objective across both stages is expressed as:

\begin{equation}
\footnotesize
\begin{split}
\mathcal{L}_{\text{MAGE-ID}} &=
\mathrm{MSE}(x_{\mathrm{num}},\hat{x}_{\mathrm{num}})
+ \mathrm{CE}(x_{\mathrm{cat}},\hat{x}_{\mathrm{cat}}) \\[-2pt]
&\quad
+ \mathrm{MSE}(x_{\mathrm{img}},\hat{x}_{\mathrm{img}})
+ \beta\, D_{\mathrm{KL}}\!\big(q_{\phi}(z|x)\,\|\,\mathcal{N}(0,I)\big) \\[-2pt]
&\quad
+ \mathbb{E}\!\left[w(\sigma)\|f_{\theta}(z+\sigma\varepsilon,\log\sigma)-z\|_2^2\right].
\end{split}
\label{eq:mageid-loss}
\end{equation}
Here, $q_{\phi}(z \mid x)$ is the VAE posterior with
$z = \mu + \sigma \odot \varepsilon$,
$\varepsilon \!\sim\! \mathcal{N}(0,I)$,
and $f_{\theta}$ is the diffusion denoiser.
$\sigma$ is the noise level and $w(\sigma)$ its EDM weight.
\noindent
The composite loss $\mathcal{L}_{\text{MAGE-ID}}$ jointly optimizes reconstruction fidelity, latent regularization, and diffusion consistency across tabular and image modalities. 
The first term, $\mathrm{MSE}\big(x_{\mathrm{num}}, \hat{x}_{\mathrm{num}}\big)$, minimizes the mean-squared reconstruction error for numerical network features, enforcing preservation of continuous statistics such as packet counts and flow durations. 
The second term, $\mathrm{CE}(x_{\mathrm{cat}}, \hat{x}_{\mathrm{cat}})$, applies a cross-entropy objective to categorical variables (e.g., protocol or flag types), guiding the model to recover discrete distributions without collapsing rare categories. 
The image reconstruction term $\mathrm{MSE}\!(x_{\mathrm{img}}, \hat{x}_{\mathrm{img}})$ constrains the CNN-based image decoder to faithfully reproduce DeepInsight feature maps, thereby aligning structural correlations between tabular and visual spaces. 
The Kullback–Leibler divergence term $D_{\mathrm{KL}}\!\big(q_{\phi}(z\mid x)\,\|\,\mathcal{N}(0,I)\big)$ regularizes the latent posterior toward an isotropic Gaussian prior, promoting smooth and disentangled latent manifolds that support reliable sampling and interpolation. 
The diffusion consistency term $\mathbb{E}\!\left[w(\sigma)\,\|f_{\theta}(z+\sigma\varepsilon,\log\sigma)-z\|_2^2\right]$ enforces noise-level-conditioned denoising accuracy within the EDM framework, where $f_{\theta}$ learns to predict the clean latent from its perturbed version. 
The weighting function $w(\sigma)$ adaptively scales this objective according to the noise magnitude, emphasizing mid-range diffusion steps that contribute most to generative stability and coverage. 
Together, these components enable MAGE-ID to reconstruct modality-specific details, maintain a coherent latent distribution, and learn a noise-robust prior that synthesizes high-fidelity, balanced intrusion data across tabular and image domains.


%% file: experiments.tex
\section{Experiments}
\label{sec:experiments}

\begin{table*}[t]
\centering
\caption{Performance Comparison of Generative Models on CIC-IDS-2017 and NSL-KDD Datasets }
\renewcommand{\arraystretch}{1.1}
\setlength{\tabcolsep}{6pt}
\begin{tabular}{lcccccc}
\toprule
\textbf{Model} & \textbf{Detectability} & \textbf{Precision} & \textbf{Recall} & \textbf{Density} & \textbf{Coverage} & \textbf{AUC (MLE)} \\
\midrule
\multicolumn{7}{c}{\textbf{CIC-IDS-2017 Dataset}} \\
\midrule
TabDDPM & 0.9356 & 0.6993 & 0.9496 & 0.5210 & 0.5957 & 0.9937 \\
TabSyn  & 0.9498 & 0.4894 & 0.9586 & 0.3140 & 0.3814 & 0.9976 \\
\textbf{MAGE-ID} & \textbf{0.9974} & \textbf{0.7562} & \textbf{0.9795} & \textbf{0.5959} & \textbf{0.7017} & \textbf{0.9997} \\
\midrule
\multicolumn{7}{c}{\textbf{NSL-KDD Dataset}} \\
\midrule
TabDDPM & 0.9020 & 0.7541 & 0.9774 & 0.5801 & 0.6264 & 0.9683 \\
TabSyn  & 0.9408 & 0.6553 & 0.9786 & 0.4877 & 0.6465 & 0.9809 \\
\textbf{MAGE-ID} & \textbf{0.9543} & \textbf{0.7978} & \textbf{0.9808} & \textbf{0.6245} & \textbf{0.6882} & \textbf{0.9993} \\
\bottomrule
\end{tabular}
\label{tab:combined_gen_metrics}
\end{table*}

\textbf{Dataset:}
We evaluate MAGE-ID on two widely used IDS benchmarks: \textit{CICIDS-2017}~\cite{sharafaldin2018toward} and \textit{NSL-KDD}~\cite{tavallaee2009detailed}.
For CICIDS-2017, we focus on \textit{PortScan} and \textit{DDoS} attacks versus \textit{Benign} traffic because these attacks are frequent in real-world incidents and exhibit distinct flow-level signatures that particularly benefit from tabular and image multimodal modeling, forming a 30{,}000-sample dataset with an 80:20 split. Each instance includes over 80 flow-level features such as packet counts, durations, and flags, offering realistic and diverse intrusion behaviors.
For NSL-KDD, a cleaned version of KDD’99, we construct a similarly imbalanced subset (80\% \textit{Normal}, 20\% \textit{Attack}) of 30{,}000 records with 41 features spanning protocol, service, and byte statistics. This dataset enables assessment of MAGE-ID’s robustness and generalization under varying network characteristics.

\textbf{DeepInsight Transformation:}
To obtain structured image representations, we use the \textit{DeepInsight} framework~\cite{sharma2019deepinsight}, which maps tabular features into a 2D layout via t-SNE–based similarity projection, producing compact $10{\times}10$ grayscale images that preserve feature correlations.

\textbf{Implementation Details:}
All models were implemented in PyTorch~2.3 and trained on a single \textbf{NVIDIA~A100~(40~GB)} GPU for \textbf{4000~epochs}. Numerical features were transformed using a Gaussian QuantileTransformer and categorical ones integer-encoded. The \textit{Tabular~VAE} employed a 3-layer Transformer (4~heads, latent~64), the \textit{Image~VAE} a 3-block CNN (32–64–128 filters, latent~64), and the \textit{Diffusion~Denoiser} an MLP~[256,256,128] trained for 1000~steps under a cosine noise schedule with AdamW (lr~$2{\times}10^{-4}$, batch~2048, weight~decay~$10^{-4}$). The KL~$\beta$ term was annealed from~1.0→0.1 over the first~30\% of epochs, and sampling used~50~EDM steps with mild stochasticity ($S_{\text{churn}}{=}3.0$).

\textbf{Evaluation Metrics:}
We assess generative quality across three dimensions: \textit{Machine Learning Efficacy (MLE)}, \textit{Detectability}, and \textit{Statistical Fidelity}. For MLE, following~\cite{zhang2024mixedtype}, a classifier (or regressor) is trained entirely on synthetic samples and evaluated on held-out real data; high \textit{Accuracy}, \textit{F1-score}, and \textit{ROC-AUC} (or \textit{$R^2$} and \textit{RMSE}) indicate strong preservation of task-relevant structure. Detectability is measured using the SDMetrics detection test~\cite{sdmetrics2024}, where a logistic-regression classifier separates real and synthetic data based on ROC–AUC; we convert this to a similarity score $\mathrm{Detectability} = 1 - 2|\mathrm{AUC} - 0.5|$ so that $1.0$ denotes indistinguishability and $0.0$ perfect separation. Statistical fidelity and diversity are evaluated using the PRDC metrics~\cite{naeem2020reliable}, where \textit{Precision} measures fidelity (staying on the real manifold), \textit{Recall} reflects coverage of the real distribution, and \textit{Density} and \textit{Coverage} quantify local concentration and the fraction of real points represented by synthetic neighbours. All metrics use the authors’ official implementations with matched sample sizes to ensure fair comparison.

\textbf{Baselines:}
We compare \textbf{MAGE-ID} with two state-of-the-art tabular diffusion models: \textbf{TabSyn}~\cite{zhang2024mixedtype} and \textbf{TabDDPM}~\cite{kotelnikov2023tabddpm}.
\textit{TabSyn} performs latent-space diffusion with a mixed-type encoder–decoder for heterogeneous data, while \textit{TabDDPM} applies denoising diffusion directly in feature space for high-fidelity synthesis. Since the tabular-only MAGE-ID follows the same latent diffusion principle as \textit{TabSyn}, it serves as our unimodal counterpart, confirming that the observed gains stem from the added image branch and cross-representation coupling. We did not include GAN- or VAE-based generators, as diffusion models (\textit{TabSyn}, \textit{TabDDPM}) outperform them and serve as stronger, complementary baselines~\cite{zhang2024mixedtype}.

\textbf{Data Splits and Evaluation Protocol:}
To avoid data leakage, each dataset was randomly divided into \textbf{70\% training}, \textbf{15\% validation}, and \textbf{15\% test} sets using a fixed seed, ensuring fully disjoint samples.
All generative models (\textit{TabSyn}, \textit{TabDDPM}, \textit{MAGE-ID}) were trained on the training set, with hyperparameters tuned on validation only.
The \textbf{MLE metric} (“train on synthetic, test on real”) used the held-out test data, which remained unseen throughout training and sampling.
No test samples were involved in diffusion, VAE optimization, or normalization, and the DeepInsight mapping was fitted solely on the training split.
This protocol ensures MLE and Detectability results reflect genuine generalization to unseen real data.
\begin{figure}[t]
    \centering
    \includegraphics[width=0.6\columnwidth]{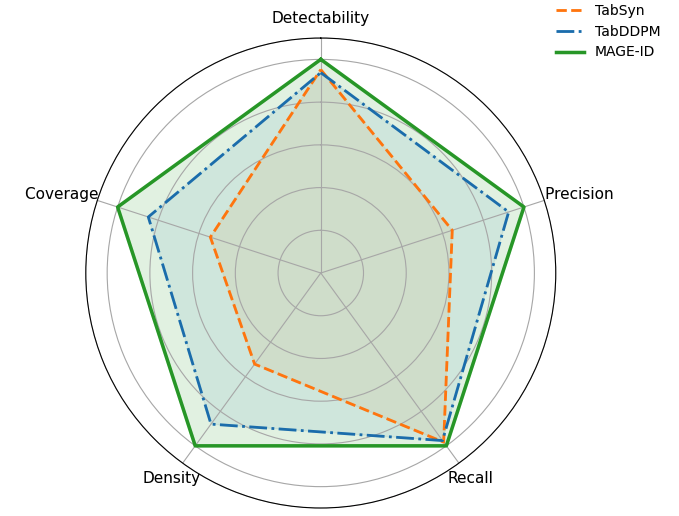}
    \caption{Radar chart comparison of TabSyn, TabDDPM, and MAGE-ID on the CIC-IDS-2017 dataset across Detectability, Precision, Recall, Density, and Coverage metrics.}
    \label{fig:cicids_radar}
\end{figure}

%% file: results.tex
\section{Results and Discussion}
\label{sec:results}

Table~\ref{tab:combined_gen_metrics} compares \textbf{MAGE-ID} against two state-of-the-art generative baselines for tabular data synthesis: 
\textbf{TabSyn}~\cite{zhang2024mixedtype} and \textbf{TabDDPM}~\cite{kotelnikov2023tabddpm}. 
These models were selected for comparison because they represent complementary diffusion-based paradigms for tabular generation.
As shown in Table~\ref{tab:combined_gen_metrics}, \textbf{MAGE-ID} achieves a consistently favorable balance between fidelity, diversity, and downstream utility across both CIC-IDS-2017~\cite{sharafaldin2018toward} and NSL-KDD~\cite{tavallaee2009detailed} datasets. 
These gains align with our design choices: (i) learning \emph{paired} tabular–image latents, which constrains off-manifold drift by enforcing structural consistency between modalities, and 
(ii) modeling the \emph{standardized multimodal} latent with a noise-conditioned diffusion denoiser, which expands generative support and mitigates mode collapse while preserving inter-modal dependencies.

\paragraph{Fidelity \& diversity (PRDC)}
Using the PRDC metrics~\cite{naeem2020reliable}, MAGE-ID yields the best \textit{Precision, Recall, Density}, and \textit{Coverage} on both datasets, indicating that samples are simultaneously realistic (on-manifold) and broadly covering the real-data support. 
On CIC-IDS-2017, MAGE-ID improves over the best baseline by \textbf{+5.69\%} in Precision (0.7562 vs.\ 0.6993), \textbf{+2.09\%} in Recall (0.9795 vs.\ 0.9586), \textbf{+7.49\%} in Density (0.5959 vs.\ 0.5210), and \textbf{+10.6\%} in Coverage (0.7017 vs.\ 0.5957).
On NSL-KDD, improvements remain consistent: \textbf{+4.37\%} Precision (0.7978 vs.\ 0.7541), \textbf{+0.22\%} Recall (0.9808 vs.\ 0.9786), \textbf{+4.44\%} Density (0.6245 vs.\ 0.5801), and \textbf{+4.17\%} Coverage (0.6882 vs.\ 0.6465).
Notably, all methods attain high PRDC \textit{Recall} ($\gtrsim 0.95$), but MAGE-ID delivers the largest gains where baselines lag most—\textit{Density} and \textit{Coverage}—which are direct indicators of improved support coverage and reduction of mode under-representation.

\paragraph{Indistinguishability and downstream utility}
The \textit{Detectability} metric from SDMetrics~\cite{sdmetrics2024} (higher is better under our setup) and the \textit{MLE AUC} corroborate the PRDC trends.
On CIC-IDS-2017, MAGE-ID reaches \textbf{0.9974} Detectability and \textbf{0.9997} MLE AUC; on NSL-KDD, \textbf{0.9543} and \textbf{0.9993}, respectively.
These outcomes indicate that classifiers trained \emph{solely} on MAGE-ID synthetic data generalize to real data with minimal loss, suggesting the model preserves task-relevant structure beyond low-level statistics. We use XGBoost for the MLE evaluation because it is a strong and widely adopted baseline that provides a stable measure of task-relevant patterns. Moreover, an XGBoost IDS (same setup as in MLE) trained on real data achieved AUCs of \textbf{0.9906} on \textit{CIC-IDS-2017} and \textbf{0.9498} on \textit{NSL-KDD}, while MAGE-ID synthetic training reached higher MLE AUCs (\textbf{0.9976} and \textbf{0.9993}), confirming that generated data effectively preserve task-relevant patterns for IDS learning. All preprocessing (normalization, encoding, and DeepInsight mapping) was fitted strictly on the training split to prevent leakage, ensuring that the near-perfect MLE AUCs reflect genuine generative fidelity rather than overfitting or data reuse.

As illustrated in Fig.~\ref{fig:cicids_radar}, the radar chart highlights that \textbf{MAGE-ID} consistently achieves superior performance across all five metrics—Detectability, Precision, Recall, Density, and Coverage—demonstrating improved fidelity and diversity compared to both \textit{TabSyn} and \textit{TabDDPM}.

\paragraph{Why multimodality helps}
Compared with unimodal baselines (TabSyn and TabDDPM), MAGE-ID’s joint latent couples tabular statistics with DeepInsight images, which encode pairwise feature relations in a 2D spatial layout~\cite{sharma2019deepinsight}. 
This cross-modal constraint raises \textit{Precision} (fidelity) by anchoring generations near the real manifold and improves \textit{Coverage/Density} by guiding the diffusion prior toward under-sampled regions (e.g., \textit{PortScan}, \textit{DDoS}), thereby addressing imbalance more effectively than tabular-only pipelines.

\paragraph{Cross-dataset robustness}
Despite differences in feature scales and attack composition, MAGE-ID shows consistent gains across both datasets, with larger margins on CIC-IDS-2017 and stable improvements on NSL-KDD, demonstrating strong generalization of its latent diffusion and $\beta$-annealed encoders. These cross-dataset improvements indicate that multimodal synthetic augmentation can enhance a model’s ability to generalize to new operating conditions, aligning with the long-term goal of developing adaptive IDS solutions.

\paragraph{Computational complexity}
In terms of runtime, \textbf{MAGE-ID} required \textbf{5600\,s} for training, compared to \textbf{5495\,s} for TabSyn and \textbf{2316\,s} for TabDDPM, with the higher cost for MAGE-ID and TabSyn attributed to their additional VAE components. Despite this, MAGE-ID offers the fastest sampling—\textbf{3.6\,s} versus \textbf{13.2\,s} (TabSyn) and \textbf{56\,s} (TabDDPM)—thanks to its latent-space diffusion design, which avoids expensive high-dimensional reconstruction. \emph{All models were trained for 4000 epochs with batch size 2048 under identical preprocessing and hardware.}

\paragraph{Takeaways for IDS augmentation}
Practically, these results suggest that (i) balanced, multimodal augmentation can increase minority-class representation without sacrificing realism; (ii) PRDC \textit{Coverage/Density} should be reported alongside \textit{Precision/Recall} to diagnose support under-coverage; and (iii) high \textit{Detectability} with strong \textit{MLE} validates that synthetic flows are both distributionally plausible and useful for training intrusion detectors.
Although the image modality in MAGE-ID is derived from tabular features, it captures spatial feature correlations that complement the original data, effectively creating a representation-level multimodal setting. This enhances cross-representation consistency and improves generative fidelity. Future work will extend MAGE-ID to truly heterogeneous modalities (e.g., PCAP, payload, and system logs) for full cross-domain dependency modeling.


%% file: conclusion.tex
\section{Conclusion}
\label{sec:Conclusion}
This work presents MAGE-ID, a unified multimodal diffusion framework for generating realistic and balanced intrusion data across tabular and image domains. By integrating DeepInsight transformations with latent diffusion modeling, MAGE-ID bridges the gap between unimodal synthesis and multimodal IDS augmentation. Experimental results show consistent gains in PRDC, Detectability, and MLE metrics across benchmark datasets, confirming improved realism and utility for training intrusion detectors. Future work will extend MAGE-ID to include temporal modalities and cross-dataset transfer learning for broader deployment in real-world cyber-defense systems.
This material is based upon work supported by the Google Cyber NYC Institutional Research Program. The views expressed are those of the authors and do not necessarily reflect those of Google.